%% file: probabilistic-gr-landmarks.tex
\def\year{2021}\relax
\documentclass[letterpaper]{article} 
\usepackage{aaai21}  
\usepackage{times}  
\usepackage{helvet} 
\usepackage{courier}  
\usepackage{url}  
\usepackage{graphicx} 
\usepackage{natbib}  
\usepackage{caption} 
\usepackage{makecell}
\usepackage{algorithm}
\usepackage[noend]{algpseudocode}

\usepackage{amsthm}
\usepackage{amsmath}
\usepackage{amssymb}
\usepackage{thmtools}
\usepackage{thm-restate}
\newtheorem{definition}{Definition}

\newcommand{\idest}{{\it i.e.}}
\newcommand{\exemp}{{\it e.g.}}

\newcommand{\etal}{{\it et al.}}

\usepackage[colorinlistoftodos, 
	]{todonotes}

\input{formalism.tex}

\title{Inferring Agents Preferences as Priors for Probabilistic Goal Recognition}
\author{
Kin Max Gusmão$^1$,
Ramon Fraga Pereira$^2$,
Felipe Meneguzzi$^3$\\
}
\affiliations{
$^1$Hewlett Packard, Brazil\\
$^2$Sapienza University of Rome, Italy\\
$^3$Pontifical Catholic University of Rio Grande do Sul, Brazil\\
gusmaokin@gmail.com,
pereira@diag.uniroma1.it
felipe.meneguzzi@pucrs.br
}
\begin{document}

\maketitle

\begin{abstract}

Recent approaches to goal recognition have leveraged planning landmarks to achieve high-accuracy with low runtime cost. 
These approaches, however, lack a probabilistic interpretation. 
Furthermore, while most probabilistic models to goal recognition assume that the recognizer has access to a prior probability representing, for example, an agent's preferences, virtually no goal recognition approach actually uses the prior in practice, simply assuming a uniform prior. 
In this paper, we provide a model to both extend landmark-based goal recognition with a probabilistic interpretation and allow the estimation of such prior probability and its usage to compute posterior probabilities after repeated interactions of observed agents. 
We empirically show that our model can not only recognize goals effectively but also successfully infer the correct prior probability distribution representing an agent's preferences. 

\end{abstract}

\section{Introduction}

\textit{Goal Recognition} is the task of inferring an agent's goals, given a potentially flawed observation of this agent's behavior~\cite{ActivityIntentPlanRecogition_Book2014}. 
The area of \textit{Goal and Plan Recognition as Planning}~\cite{RamirezG_IJCAI2009} has advanced substantially over the past decade, yielding a number of approaches capable of coping with partial and noisy observations~\cite{NASA_GoalRecognition_IJCAI2015,Sohrabi_IJCAI2016}, and doing this efficiently~\cite{Pereira2020:AIJ}.

Virtually, all such efforts use the model of \citeauthor{RamirezG_AAAI2010}~\shortcite{RamirezG_AAAI2010} as their underpinning, which defines via Bayes' Rule the probability of a goal, given observations in terms of the probability of the observations given the goal, and some prior probability of goals, representing an agent's preference. 
Comparatively, fewer efforts provide a probabilistic interpretation of the model defined by \citeauthor{RamirezG_AAAI2010}~\cite{Sohrabi_IJCAI2016,GalMor_AAAI2018}.
Fewer efforts still actually use the prior probability on goals, assuming instead a uniform distribution for the goals, and ignoring the prior in their calculations. 
Ignoring the prior probability bakes into the goal recognition model the assumption that all goal recognition tasks are \textit{one-shot}, such that agents pursue exactly one goal within a particular goal recognition domain exactly once. 
Such an assumption does not reflect many goal recognition tasks, such as intention recognition for elder care~\cite{ProblemsWithElderCare_AAAI2002}, assistance for activities of daily living~\cite{DailyPR_2010}, proactive user interfaces~\cite{AmirGal2013}, among others. 

In this paper, we expand recognition problems from the traditional \textit{one-shot} setting used by all models so far into problems that assume goal hypotheses have different probability distributions representing an agent's preferences and develop a solution for this problem by extending recent work on landmark-based goal recognition~\cite{Pereira2020:AIJ}. 
Our key contributions are twofold: (1) a novel definition of goal recognition problems with a goal preference distribution; and (2) a probabilistic interpretation that relies on the concept of landmarks. 

\section{Background}

\noindent \textbf{Planning.} \textit{Planning} is the problem of finding a sequence of actions (\idest, a plan) that achieves a goal state from an initial state~\cite{AutomatedPlanning_Book2011}.
A \textit{state} is a finite set of facts that represent logical values according to some interpretation. 
\textit{Facts} can be either positive or negated ground predicates.
A predicate is denoted by an n-ary predicate symbol $p$ applied to a sequence of zero or more terms ($\tau_0$, $\tau_1$, ..., $\tau_n$). 
An \textit{operator} is represented by a triple $\action = \langle \name(\action), \pre(\action), \eff(\action) \rangle$ where $\name(\action)$ represents the description or signature of $\action$; $\pre(\action)$ describes the preconditions of $\action$ --- a set of facts or predicates that must exist in the current state for $\action$ to be executed; $\eff(\action)=\eff(\action)^+ \cup \eff(\action)^-$ represents the effects of $\action$, with $\eff(\action)^+$  an \textit{add-list} of positive facts or predicates, and $\eff(\action)^-$ a \textit{delete-list} of negative facts or predicates. 
When we instantiate an operator over its free variables, we call the resulting ground operator an \emph{action}. 
A \textit{planning instance} is represented by a triple $\planningtask = \langle \planningdomain, \initialstate, \goalcondition\rangle$, where $\planningdomain = \langle\fluents,\actions\rangle$ is a \textit{planning domain definition}; $\fluents$ consists of a finite set of facts and $\actions$ a finite set of actions; $\initialstate$ $\subseteq$ $\fluents$ is the initial state; and $G$ $\subseteq$ $\fluents$ is the goal state. 
A \textit{plan} is a sequence of actions $\plan = \langle \action_0, \action_1, ..., \action_n \rangle$ that modifies the initial state $\initialstate$ into one in which the goal state $G$ holds by the successive execution of actions in $\plan$. As in \textit{Classical Planning}, actions have an associated cost, and here, we assume that this cost is 1 for all actions. A plan $\plan$ is considered \textit{optimal} if its cost, and thus length, is minimal.

\noindent \textbf{Goal Recognition as Planning.} \textit{Goal Recognition} is the task of discerning the intended goal of autonomous agents or humans by observing their interactions in a particular environment~\cite[Chapter 1]{ActivityIntentPlanRecogition_Book2014}. 
We formally define the problem of \textit{Goal Recognition as Planning} by adopting the formalism proposed by Ram{\'{\i}}rez and Geffner~\shortcite{RamirezG_IJCAI2009,RamirezG_AAAI2010}, as follows in Definition~\ref{def:GoalRecognition}. 

\begin{definition}[\textbf{Goal Recognition Problem}]\label{def:GoalRecognition}
A goal recognition problem is a tuple $\grtask = \langle\planningdomain,\initialstate ,\goalconditions, \observations\rangle$, where: $\planningdomain = \langle\fluents, \actions\rangle$ is a planning domain definition; $\initialstate$ is the initial state; $\goalconditions = \langle G_0, G_1, ..., G_n \rangle$ is the set of goal hypothesis, including the correct intended goal $G^{*}$, such that $G^{*} \in \goalconditions$; and $\observations = \langle o_0, o_1, ..., o_n \rangle$ is an observation sequence of executed actions, with each observation $o_i \in \actions$.
\end{definition}

The ideal solution for a goal recognition problem $\grtask$ is the correct intended goal $G^{*} \in \goalconditions$ that the observation sequence $\observations$ of a plan execution achieves. 
An observation sequence $\observations$ can be \textit{full} or \textit{partial}. A \textit{full observation sequence} contains all actions of agents' plans, so all actions of a plan are observed, whereas in a \textit{partial observation sequence}, only a sub-sequence of actions are observed. 

Existing work on \textit{Goal Recognition as Planning} considers the solution to a goal recognition problem to be either a \textit{score system} associated to the set of goal hypothesis~\cite{RamonNirMeneguzzi_AAAI2017,PereiraPM19,Pereira2020:AIJ}, or a \textit{probability distribution} for the goal hypothesis~\cite{RamirezG_IJCAI2009,RamirezG_AAAI2010,NASA_GoalRecognition_IJCAI2015,Sohrabi_IJCAI2016,Pereira:2019em}. 
In this work, we extend a landmark-based approach for goal recognition and provide a probabilistic model that relies on the concept of \textit{landmarks}.


\section{Probabilistic Goal Recognition as Reasoning over Landmarks} \label{sec:prob-gr}

Key to our probabilistic goal recognition approach is the concept of landmarks in planning, which has been extensively used in goal recognition approaches~\cite{PereiraMeneguzzi_ECAI2016,RamonNirMeneguzzi_AAAI2017,MorEtAl_AAMAS18,PozancoCounterPlanningEFB18,ShvoM_AAAI20}. 
\textit{Landmarks} are defined as necessary fact (or actions) that must be true (or executed) at some point along all valid plans that achieve a particular goal from an initial state~\cite{Hoffmann2004_OrderedLandmarks}. 
Landmarks are often partially ordered based on the sequence in which they must be achieved. 
Hoffman \etal~\shortcite{Hoffmann2004_OrderedLandmarks} define fact landmarks as follows:

\begin{definition}[\textbf{Fact Landmark}]\label{def:planLandmark}
Given a planning instance $\planningtask = \langle \planningdomain, \initialstate, G\rangle$, a formula $L$ is a fact landmark in $\planningtask$ iff $L$ is true at some point along all valid plans that achieve $G$ from $\initialstate$. 
A landmark is a type of formula (\exemp, a conjunctive or disjunctive formula) over a set of facts that must be satisfied at some point along all valid plan executions.
\end{definition}

The process of generating all landmarks and deciding their ordering is proved to be PSPACE-complete~\cite{Hoffmann2004_OrderedLandmarks}, which is exactly the same complexity as deciding plan existence~\cite{PlanningComplexity_Bylander1994}. 
Thus, to operate efficiently, most landmark extraction algorithms~\cite{Hoffmann2004_OrderedLandmarks,RHW_AAAI_2008,HM_ECAI_2010} extract only a subset of landmarks for a given planning instance. 

In what follows, we expand the \textit{landmark-based goal recognition framework} of~\citeauthor{Pereira2020:AIJ}~\shortcite{Pereira2020:AIJ} by introducing a probabilistic interpretation that allows us to perform recognition repeatedly refining estimated goal probabilities over time. 
The recognition framework of~\citeauthor{Pereira2020:AIJ}~\shortcite{Pereira2020:AIJ} provides a \textit{score system} that ranks the goal hypothesis $\goalconditions$ according to the ratio between the \textit{achieved landmarks} and \textit{total number of landmarks}.
Our probabilistic interpretation model is based the well-known probabilistic model of~\citeauthor{RamirezG_AAAI2010}~\shortcite{RamirezG_AAAI2010}. The probabilistic model of \cite{RamirezG_AAAI2010} sets the probability distribution for every goal $\goalcondition$ in the set of goals $\goalconditions$, and the observation sequence $\observations$ to be a Bayesian posterior conditional probability, as follows:
\begin{equation}
\label{eq:rg_posterior}
\probability{G \mid \observations} = \alpha * \probability{\observations \mid G} * \probability{G}
\end{equation}
\noindent where $\probability{G}$ is a \textit{prior probability} to goal $G$, $\alpha$ is a \textit{normalizing factor}, and $\probability{\observations \mid G}$ is the probability of observing $\observations$ when the goal is $G$. \citeauthor{RamirezG_AAAI2010}~\shortcite{RamirezG_AAAI2010} compute $\probability{\observations \mid G}$ by computing two plans for every goal $G$, and based on these two plans, they compute a \textit{cost-difference} between these plans and plug it into a Boltzmann equation. Basically, they compute a plan that \textit{complies} with the observations, and another a plan that \textit{does not comply} with the observations.
The intuition of \citeauthor{RamirezG_AAAI2010}~probabilistic model is that the lower the \textit{cost-difference} for a goal, the higher the probability for this goal.

In contrast, our probabilistic model reasons over the evidence of landmarks, and follows the intuition of~\citeauthor{Pereira2020:AIJ}~\shortcite{Pereira2020:AIJ}, where goals $\goalconditions$ are ranked according to their score, namely, the most likely goals are the ones that have achieved most of their landmarks in the observations.
Thus, replicating this ranking in a probabilistic setting entails assigning probabilities to the observation of landmarks. 
If we consider an arbitrary goal $G$ and represent its landmarks as a set $\mathcal{L}_G$, where $L_G \in \mathcal{L}_G$ is an individual landmark for $G$, we can reason about the probabilistic properties of observing such landmarks. 
First, since landmarks are \textit{necessary conditions} to achieve a goal, the probability of observing all landmarks in a set of observations for a given goal should be $1$, as we formally define in Equation~\ref{eq:prob_all_landmarks}.

\begin{align}
    \label{eq:prob_all_landmarks}
    \probability{\mathcal{L}_G \mid G} & = \sum_{L_G \in \mathcal{L}_G} \probability{L_G \mid G} = 1 
\end{align}

Without any additional evidence, we can also infer that the probability of observing any given individual landmark in an observation sequence $\observations$ should be uniformly distributed as shown in Equation~\ref{eq:prob_one_landmark}. 

\begin{align}
    \label{eq:prob_one_landmark}
    \probability{L_G \mid G} & = \frac{1}{|\mathcal{L}_G|}
\end{align}    

If we completely ignore the ordering of the landmarks in observations, and consider only the probabilities of observing landmarks, we can compute the probability of a particular set of observations $\observations$ towards a goal $G$ using Equation~\ref{eq:pog}.

\begin{align}
    \label{eq:pog}
    \probability{\observations \mid G} & = \displaystyle \sum_{L_G \in (\mathcal{L}_G \cap \observations)} \probability{L_G \mid G}
\end{align} 
 
Thus, we use landmarks as a proxy for the probability of the entire set of observations $\observations$ given a goal $G$. 
We can plug $\probability{\observations \mid G}$ defined in Equation~\ref{eq:pog} into the Bayesian formulation of \citeauthor{RamirezG_AAAI2010}~from Equation~\ref{eq:rg_posterior}.
Since we assume the set of goal hypotheses to be exhaustive and mutually exclusive, we can compute instead a \textit{normalizing factor} $\alpha$, which we obtain from Equation~\ref{eq:alpha}. 

\begin{align}
    \label{eq:alpha}
    \alpha & = \frac{1}{\displaystyle\sum_{G \in \goalconditions} \probability{\observations \mid G} * \probability{G}}
\end{align}
%


%
When no priors $\probability{G}$ are informed, we can assume that their distribution is uniform, and compute them through $\probability{G} = \frac{1}{|\goalconditions|}$. In Section~\ref{sec:tendency}, we show how we infer prior probabilities by observing repeated goal recognition episodes.

\section{Prior Estimation by Repeated Episodes} \label{sec:tendency}

We now expand the probabilistic model of Section~\ref{sec:prob-gr} to compute posterior goal probabilities when the prior goal probabilities follow a \textit{non-uniform} distribution over repeated goal-recognition episodes. 
The resulting model allows us to converge towards the actual probability distribution that can be used as a prior for further goal recognition episodes. 
We formalize the extended version of such problem in Definition~\ref{def:seqPlanRecognition}. 

\begin{definition}[\textbf{Repeated Goal Recognition Problem}]\label{def:seqPlanRecognition}
A repeated goal recognition problem is a tuple $\repgrtask = \langle\planningdomain, \initialstate, \goalconditions, \seqobservations\rangle$, where: $\planningdomain = \langle\fluents, \actions\rangle$ is a planning domain definition; $\initialstate$ is the initial state; 
$\goalconditions = \langle \goalcondition_0, \goalcondition_1, ..., \goalcondition_n \rangle$ is the set of goal hypothesis; and 
$\seqobservations = \set{\observations_0, \dots, \observations_n}$ is a set of observation sequences, where each
$\observations_{i} \in \seqobservations$ is an observation sequence $\langle \observation_0, \observation_1, \dots, \observation_m \rangle$  of executed actions, with each observation $\observation_i \in \actions$. 
Observation sequences $\observations_{i}$ are projections of plans $\plan_{i}$ for planning tasks $\tuple{\planningdomain, \initialstate, \goalcondition_{i}}$ such that the intended goal 
$\goalcondition_{i} \in \goalconditions$ is drawn from a probability distribution 
$\probability{ \goalconditions }$ 
with probability 
$\probability{\goalconditions = \goalcondition_{i}}$.
\end{definition}

The solution for a repeated goal recognition problem is the correct probability distribution $\probability{ \goalconditions }$  that generated the set of observation sequences $\observations$ in the problem of Definition~\ref{def:seqPlanRecognition}. 
Here, $\probability{ \goalconditions }$ does not represent the result of a single episode of goal recognition, but rather the goal preferences of the agent under observation under repeated episodes. 


Our prior estimation consists of processing each observation sequence $\observations_{i} \in \seqobservations$ and count the number of times we recognize each candidate goal as the actual goal of an observation sequence $\observations_{i}$. 
We recognize the goals of each observation sequence independently, ignoring any priors in order to avoid biasing the count (Line~\ref{alg:line:recognize}).
After each run, we check whether we correctly recognize the goal for sample $\observations_{i}$ (Line~\ref{alg:line:increaseCount}), which we do in a supervised way. 
Each correctly recognized goal $\goalcondition$ for a sample results in an increment of the corresponding counter $\goalcounter_{\goalcondition}$. 
After repeating the process for all samples, we compute the prior for every candidate using the counter values and a form of \textit{Laplace smoothing}~\cite{marquis1825essai} shown in Line~\ref{alg:line:laplace}, where $k$ is the number of ghost samples we include to prevent any goal from having a probability of exactly $0$. 
Algorithm~\ref{alg:prior_estimation} formally describes how our prior estimation process works.

\begin{algorithm}[tbh]
\caption{Prior Estimation.}
\label{alg:prior_estimation}
\begin{algorithmic}[1]
\small
\Function{estimatePrior}{$\repgrtask$}
    \State $\goalcounter_{\goalcondition} \gets 0$ for all $\goalcondition \in \goalconditions$ \label{line:init} 
    \For{$\observations \in \seqobservations$ }\label{line:candidateEstimationStart}
		\State{$\mathbf{\goalcondition} \gets \Call{recognize}{\grtask}$}\label{alg:line:recognize}
        \If{$\grsolution \in \mathbf{\goalcondition}$} \label{alg:line:increaseCount}
            $\goalcounter_{\goalcondition} \gets \goalcounter_{\goalcondition} +1$ for all $\goalcondition \in \mathbf{\goalcondition}$
        \EndIf
    \EndFor\label{line:candidateEstimationEnd}
	\State{$\probability{\goalcondition} \gets \frac{k+\goalcounter_{\goalcondition}}{ (k*|\goalconditions|)+\displaystyle\sum_{\goalcondition \in \goalconditions}\goalcounter_{\goalcondition}}$ for all $\goalcondition \in \goalconditions$} \label{alg:line:laplace} 
    \State \textbf{return} $\probability{\goalconditions}$ 
    \Comment{Return probability distribution.} 
    \label{line:endNextStates2}
\EndFunction
\end{algorithmic}
\end{algorithm}




\section{Experiments and Evaluation} \label{sec:experiments}

We empirically evaluate our probabilistic model over the recognition datasets from \cite{RamirezG_IJCAI2009}. 
These datasets comprise hundreds recognition problems for four planning domains (\textsc{Blocks-World}, \textsc{Easy-IPC-Grid}, \textsc{Intrusion-Detection}, and \textsc{Logistics}), having recognition problems with both partial and full observability. Recognition problems with partial observability have four observation levels: 10\%, 30\%, 50\% and 70\%.

\input{table_results.tex}


\subsection{Repeated Goal Recognition Setup}


To evaluate our repeated goal recognition algorithm, we develop a recognition problem generator that generates a set of \textit{samples} that comprises $\seqobservations$ from a set of possible goal hypothesis $\goalconditions$. 
Essentially, we produce a number of planning tasks\footnote{To generate the \textit{samples}, we compose planning tasks based on initial states $\initialstate$ and goals hypothesis $\goalconditions$ of the recognition problems in the datasets from \cite{RamirezG_IJCAI2009}.} $\planningtask_{i} = \langle \planningdomain, \initialstate, \goalcondition_{i}\rangle$, such that the solution for each $\planningtask_{i}$ is a plan $\plan_{i}$ from which we generate observations $\observations_{i}$ subject to the desired level of observability, including it in $\seqobservations$. 
We use Fast Downward~\cite{HelmertFastDownward_2011} to generate the plans from which we project the observations. 
The goal state $\goalcondition_{i} \in \goalconditions$, and $\goalcondition$ is drawn from a probability distribution $\probability{ \goalconditions }$. 
For our experiments, we generate $10 * |\goalconditions|$ \textit{samples} per repeated goal recognition problem.
The probability distribution $\probability{ \goalconditions }$ is known only to the generator.
%
%
We use two different probability distributions to generate such samples: a \textit{normal} distribution with $\mu=1$ and $\sigma=0$, which we denote as \textsc{Normal-Single} distribution, where all samples have the same goal state; and a \textit{normal} distribution, such that a single (preferred) goal $\goalcondition_{i}$ has $\probability{\goalcondition_{i}}=0.5$, and the probabilities for other candidates follow a normal distribution with goals more similar to $\goalcondition_{i}$ have higher probability, resulting in a distribution with $\mu\approx1.7$ and $\sigma\approx2.4$. We denote this second probability distribution as \textsc{Normal-Diverse}.
%
After all samples have been generated, we perform the \textit{smoothing} process from Section~\ref{sec:tendency}, to smooth out the distribution using $k=1$. 


\subsection{Evaluation Metrics}

We use three metrics in our evaluation: \textit{Accuracy} (\textbf{Acc} \%), representing the fraction of problems in which the correct intended goal is among the goals with the highest posterior probability; \textit{Spread in} $\goalconditions$ (\textbf{S in} $\goalconditions$), representing the average number of goals recognized as the most likely; and \textit{recognition time} (\textbf{Time}) in seconds, representing the recognition time including the landmark extraction process.

We use two additional metrics when evaluating our probabilistic model with prior probabilities. 
\textbf{Max-Norm} is the largest difference between corresponding probabilities in the distribution that generated the samples and the estimated distribution of priors, used to evaluate the distance between these two distributions. 
If we can infer the priors exactly right, $\textbf{Max-Norm}=0$. 
The second metric is a $\Delta$ metric, which is the difference between the \probability{G \mid \observations} of the real goal when using priors and when not using priors and gives us an insight on how helpful the priors are in one-shot recognition.


\subsection{Goal Recognition Results}

Table~\ref{tab:results} shows the results for executions with \textbf{no priors} (traditional \textit{one-shot} recognition, denoted as \textsc{No Priors}), with priors generated through a \textbf{single-goal} samples distribution (\textsc{Normal-Single}), and with priors generated through \textbf{normal} samples distribution (\textsc{Normal-Diverse}). 
We show the results for all four domains using the recognition datasets from \cite{RamirezG_IJCAI2009}. For each domain, we show the number of problems (under the domain name), the average number of candidate goals $|\goalconditions|$, the average number of extracted landmarks $|\landmarks|$, and the average number of observations $|\observations|$.
For each of the three prior setups, we show recognition time, accuracy, and Spread in $\goalconditions$. 
As for the prior setups that use priors, we show results for two additional metrics: \textbf{Max-Norm} and $\Delta$.
We can see that when using \textbf{no priors} we achieve similar results (in terms of accuracy and Spread in $\goalconditions$) to the landmark-based approaches in~\cite{Pereira2020:AIJ}. 
However, we achieve much better results when using prior probabilities (\textsc{Normal-Single} and \textsc{Normal-Diverse} columns in Table~\ref{tab:results}), as it simulates agents' preference using our prior estimation process. 
Naturally, the \textsc{Normal-Single} distribution yields better results, as the agent always chooses the same intended goal in the samples.


Finally, we see that the average \textbf{Max-Norm} value is relatively high for both the \textsc{Normal-Single} and \textsc{Normal-Diverse} distributions, especially for \textsc{Normal-Diverse} distribution (on average). 
This discrepancy is likely due to the small number of samples relative to the number of goal hypotheses since there might not have been enough opportunities in the samples to count all goals with non-zero probability in the distribution. 
We expect this metric to drop when dealing with a higher number of samples. 
The $\Delta$ metric increases with the observability level.  
As the accuracy increases with the increase in observations, the probabilistic model is correct more often during the prior estimation process, which helps to increase the probability of the correct intended goal in the prior. 

\section{Conclusions}

In this paper, we have developed a novel probabilistic model for \textit{Goal Recognition as Planning} that relies on the concept of landmarks, and a prior estimation process that infers prior probabilities from past recognition episodes. 
We have shown that our probabilistic model clearly benefits when using prior probabilities that have been inferred from past recognition episodes.

Our landmark-based probabilistic model can be used not only in \textit{Classical Planning} settings, but also in other planning settings that define the concept of landmarks, i.e., \textit{Temporal Planning} landmarks~\cite{KarpasWWH15_TemporaLandmarks}, \textit{Numeric Planning} landmarks~\cite{ScalaHMT17_Landmarks}. 
Our prior estimation mechanism is completely independent of the underlying goal recognition algorithm, and any such algorithm (even a non-probabilistic one) could be used in estimating the priors.

As future work, we intend to expand our prior estimation algorithm to non-classical planning settings, as well as to settings where the agent under observation is adversarial, for example, deliberately choosing undesired goals to skew the prior probability away from the preference relation. 



\bibliography{probabilistic-gr-landmarks}

\end{document}

%% file: formalism.tex
\DeclareMathOperator{\pre}{pre}
\DeclareMathOperator{\eff}{eff}
\DeclareMathOperator{\name}{name}

\providecommand\tuple[1]{\ensuremath{\langle{#1}\rangle}}
\providecommand\set[1]{\ensuremath{\left\{ {#1} \right\}}}

\providecommand\probability[1]{\ensuremath{\mathbb{P}\left[{#1}\right]}}

\DeclareMathOperator{\fluents}{\mathcal{F}}
\DeclareMathOperator{\actions}{\mathcal{A}}

\DeclareMathOperator{\action}{a}

\DeclareMathOperator{\landmarks}{\mathcal{L}}
\providecommand{\goalcounter}{\ensuremath{\mathcal{C}}}

\DeclareMathOperator{\planningtask}{\Pi}
\DeclareMathOperator{\planningdomain}{\Xi}

\providecommand{\goalconditions}{\ensuremath{\mathcal{G}}}
\DeclareMathOperator{\plan}{\pi}

\providecommand\initialstate{\ensuremath{\mathcal{I}}}
\providecommand\goalcondition{\ensuremath{G}}

\DeclareMathOperator{\grtask}{\Pi_{\goalconditions}^{\observations}}
\DeclareMathOperator{\repgrtask}{\Pi^{\circlearrowright}_{\goalconditions}}

\providecommand{\grsolution}{\ensuremath{\goalcondition^{*}}}

\DeclareMathOperator{\observations}{\Omega}

\DeclareMathOperator{\observation}{o}
\DeclareMathOperator{\seqobservations}{\Omega^{\goalconditions}}



%% file: table_results.tex
\begin{table*}[th!]
\centering
\fontsize{7}{8.5}\selectfont
\setlength\tabcolsep{2pt}
\begin{tabular}{|c|c|c|c c|c|c|c|c|c|c|c|c|c|c|c|c|c|}
    \hline
    
    \multicolumn{5}{|c|}{} & \multicolumn{3}{c|}{\textsc{No Priors}} & \multicolumn{5}{c|}{\textsc{Normal-Single}} & \multicolumn{5}{c|}{\textsc{Normal-Diverse}}\\
    
    \hline
    
    \textbf{Domain} & $|\goalconditions|$ & $|\mathcal{L}|$ & \textbf{\% Obs} & $\observations$ & \textbf{Time} & \textbf{Acc \%} & \textbf{S in $\goalconditions$} & \textbf{Time} & \textbf{Acc \%} & \textbf{S in $\goalconditions$} & \textbf{Max-Norm} & $\Delta$ & \textbf{Time} & \textbf{Acc \%} & \textbf{S in $\goalconditions$} & \textbf{Max-Norm} & $\Delta$ \\
    
    \hline 

\makecell{\sc Blocks-World \\ (793)} & 20.3 & 12.0 & \makecell{10\\30\\50\\70\\100} & \makecell{1.1\\2.9\\4.3\\6.4\\8.6} & \makecell{0.230\\0.352\\0.346\\0.174\\0.358} & \makecell{21.9\%\\39.3\%\\59.0\%\\80.9\%\\100.0\%} & \makecell{1.3\\1.2\\1.2\\1.2\\1.5} & \makecell{0.153\\0.157\\0.164\\0.169\\0.176} & \makecell{67.9\%\\96.5\%\\96.7\%\\97.8\%\\100.0\%} & \makecell{1.2\\1.0\\1.0\\1.0\\1.5} & \makecell{0.610\\0.363\\0.280\\0.229\\0.257} & \makecell{0.350\\0.607\\0.683\\0.721\\0.687} & \makecell{0.155\\0.161\\0.166\\0.173\\0.185} & \makecell{43.9\%\\83.6\%\\90.0\%\\86.9\%\\65.6\%} & \makecell{1.1\\1.0\\1.0\\1.0\\1.0} & \makecell{0.315\\0.199\\0.154\\0.130\\0.165} & \makecell{0.184\\0.313\\0.366\\0.392\\0.360} \\ 

\hline 

\makecell{\sc Easy-Ipc-Grid \\ (390)} & 8.3 & 6.8 & \makecell{10\\30\\50\\70\\100} & \makecell{1.8\\4.4\\7.0\\9.8\\13.4} & \makecell{0.413\\0.474\\0.637\\0.379\\0.438} & \makecell{71.1\%\\86.7\%\\96.7\%\\98.9\%\\100.0\%} & \makecell{2.7\\1.6\\1.2\\1.0\\1.0} & \makecell{0.618\\0.637\\0.640\\0.632\\0.655} & \makecell{98.9\%\\97.8\%\\100.0\%\\100.0\%\\100.0\%} & \makecell{1.1\\1.0\\1.0\\1.0\\1.0} & \makecell{0.382\\0.255\\0.194\\0.132\\0.079} & \makecell{0.456\\0.559\\0.596\\0.623\\0.644} & \makecell{0.609\\0.622\\0.609\\0.644\\0.655} & \makecell{73.3\%\\97.0\%\\100.0\%\\99.6\%\\100.0\%} & \makecell{1.0\\1.0\\1.0\\1.0\\1.0} & \makecell{0.246\\0.156\\0.112\\0.084\\0.063} & \makecell{0.137\\0.273\\0.336\\0.369\\0.399} \\ 

\hline 

\makecell{\sc Intrusion-Detection \\ (390)} & 16.7 & 13.8 & \makecell{10\\30\\50\\70\\100} & \makecell{1.9\\4.5\\6.7\\9.5\\13.1} & \makecell{0.478\\0.491\\0.467\\0.460\\0.524} & \makecell{75.6\%\\94.4\%\\100.0\%\\100.0\%\\100.0\%} & \makecell{1.4\\1.0\\1.0\\1.0\\1.0} & \makecell{0.330\\0.337\\0.346\\0.385\\0.360} & \makecell{100.0\%\\100.0\%\\100.0\%\\100.0\%\\100.0\%} & \makecell{1.0\\1.0\\1.0\\1.0\\1.0} & \makecell{0.293\\0.107\\0.086\\0.085\\0.085} & \makecell{0.587\\0.727\\0.739\\0.739\\0.729} & \makecell{0.336\\0.334\\0.344\\0.404\\0.367} & \makecell{100.0\%\\100.0\%\\100.0\%\\100.0\%\\100.0\%} & \makecell{1.0\\1.0\\1.0\\1.0\\1.0} & \makecell{0.148\\0.060\\0.054\\0.053\\0.051} & \makecell{0.357\\0.458\\0.459\\0.461\\0.463} \\ 

\hline 

\makecell{\sc Logistics \\ (390)} & 10.0 & 14.3 & \makecell{10\\30\\50\\70\\100} & \makecell{2.0\\5.9\\9.6\\13.5\\18.7} & \makecell{0.544\\0.666\\0.701\\0.459\\0.675} & \makecell{62.2\%\\86.7\%\\94.4\%\\97.8\%\\100.0\%} & \makecell{2.0\\1.3\\1.1\\1.0\\1.0} & \makecell{0.518\\0.532\\0.543\\0.551\\0.590} & \makecell{100.0\%\\100.0\%\\100.0\%\\100.0\%\\100.0\%} & \makecell{1.0\\1.0\\1.0\\1.0\\1.0} & \makecell{0.513\\0.281\\0.168\\0.115\\0.082} & \makecell{0.391\\0.606\\0.694\\0.732\\0.755} & \makecell{0.530\\0.553\\0.557\\0.572\\0.607} & \makecell{79.6\%\\99.6\%\\100.0\%\\100.0\%\\100.0\%} & \makecell{1.0\\1.0\\1.0\\1.0\\1.0} & \makecell{0.262\\0.143\\0.092\\0.067\\0.060} & \makecell{0.150\\0.311\\0.379\\0.417\\0.439} \\ 

\hline 

\end{tabular}
    
\caption{Experimental results comparing our landmark-based probabilistic model with \textit{no prior} probability distribution, \textit{normal single-goal} probability distribution, and \textit{normal} probability distribution.}
\label{tab:results}
\end{table*}